\documentclass[sigconf]{acmart}
\renewcommand\footnotetextcopyrightpermission[1]{}

\usepackage{listings}
\usepackage{algorithm}
\usepackage{algorithmic}
\usepackage{stfloats}
\AtBeginDocument{%
  }
    
\copyrightyear{}
\acmYear{}
\acmConference{}
\acmBooktitle{}

\begin{document}

\title{EasyAnimate: High-Performance Video Generation Framework with Hybrid Windows Attention and Reward Backpropagation}

\author{Jiaqi Xu}
\authornote{Both authors contributed equally to this research.}
\email{zhoumo.xjq@alibaba-inc.com}
\author{Kunzhe Huang}
\authornotemark[1]
\email{huangkunzhe.hkz@alibaba-inc.com}
\author{Xinyi Zou}
\email{zouxinyi.zxy@alibaba-inc.com}
\affiliation{%
\institution{Alibaba Cloud}
\city{Hangzhou}
\country{China}}
\author{Yunkuo Chen}
\email{chenyunkuo.cyk@alibaba-inc.com}
\author{Bo Liu}
\email{xuanyuan.lb@alibaba-inc.com}
\author{Mengli Cheng}
\email{mengli.cml@alibaba-inc.com}
\affiliation{%
\institution{Alibaba Cloud}
\city{Hangzhou}
\country{China}}
\author{Jun Huang}
\email{huangjun.hj@alibaba-inc.com}
\author{Xing Shi}
\email{shubao.sx@alibaba-inc.com}
\affiliation{%
\institution{Alibaba Cloud}
\city{Hangzhou}
\country{China}}


\begin{abstract}
This paper introduces EasyAnimate, an efficient and high quality video generation framework that leverages diffusion transformers to achieve high-quality video production, encompassing data processing, model training, and end-to-end inference. 
Despite substantial advancements achieved by video diffusion models, existing video generation models still struggles with slow generation speeds and less-than-ideal video quality.
To improve training and inference efficiency without compromising performance, we propose Hybrid Window Attention. 
We design the multidirectional sliding window attention in Hybrid Window Attention, which provides stronger receptive capabilities in 3D dimensions compared to naive one, while reducing the model's computational complexity as the video sequence length increases.
To enhance video generation quality, we optimize EasyAnimate using reward backpropagation to better align with human preferences.
As a post-training method, it greatly enhances the model's performance while ensuring efficiency. 
In addition to the aforementioned improvements, EasyAnimate integrates a series of further refinements that significantly improve both computational efficiency and model performance.
We introduce a new training strategy called Training with Token Length to resolve uneven GPU utilization in training videos of varying resolutions and lengths, thereby enhancing efficiency. 
Additionally, we use a multimodal large language model as the text encoder to improve text comprehension of the model.
Experiments demonstrate significant enhancements resulting from the above improvements.
The EasyAnimate achieves state-of-the-art performance on both the VBench leaderboard and human evaluation. Code and pre-trained models are available at https://github.com/aigc-apps/EasyAnimate.

\begin{figure*}[h]
	\centering
	\includegraphics[width=0.87\linewidth]{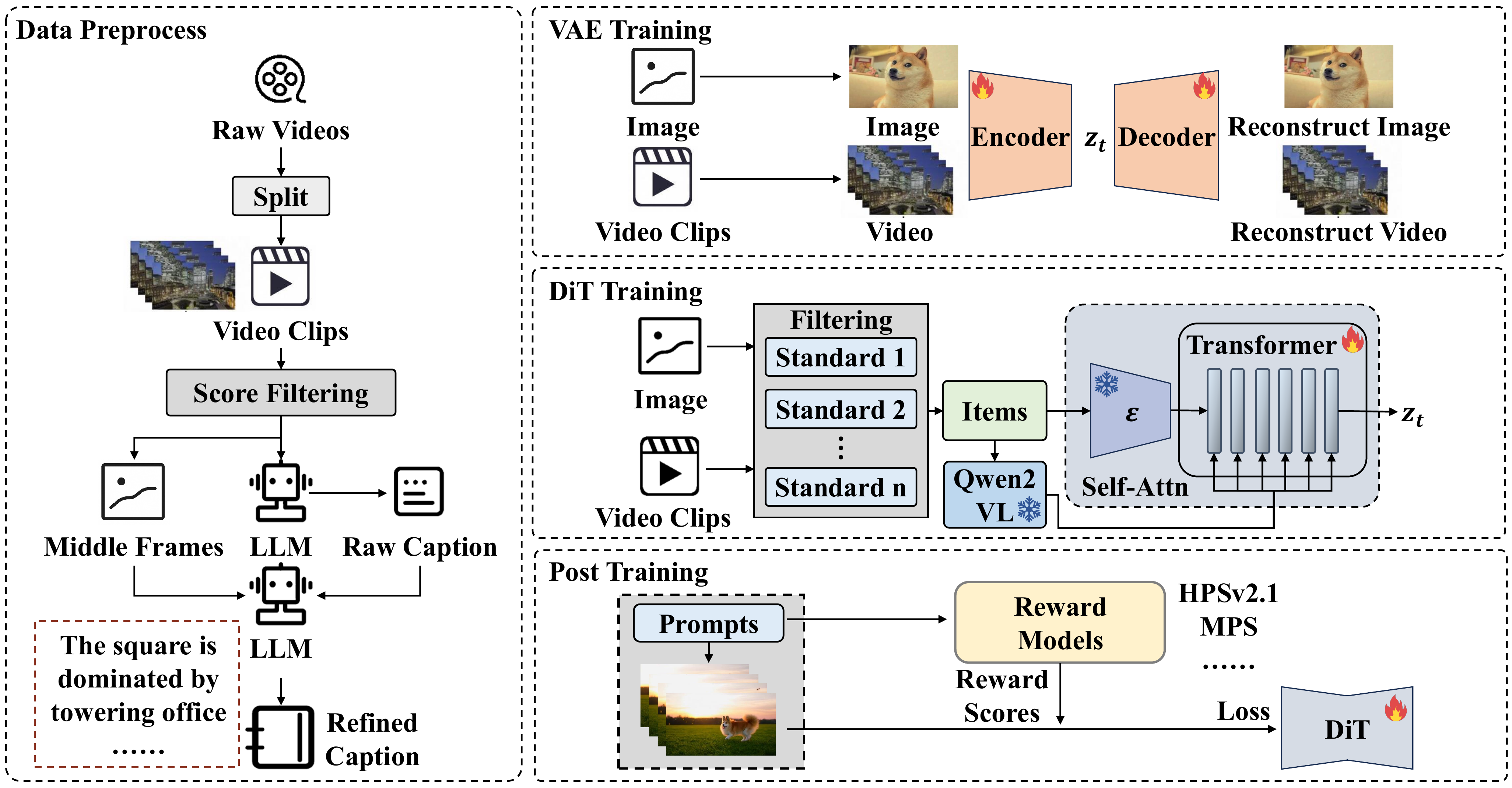}
	\caption{The EasyAnimate pipeline comprises four stages: data preprocessing, VAE Training, DiT Training and Post Training. }\label{fig:pipeline}
\end{figure*}

\end{abstract}

\ccsdesc[500]{Computing methodologies~Computer vision}

\keywords{Video Generation, Diffusion Transformers, Foundation Models}
\begin{teaserfigure}
  \centering
  \includegraphics[width=0.98\textwidth]{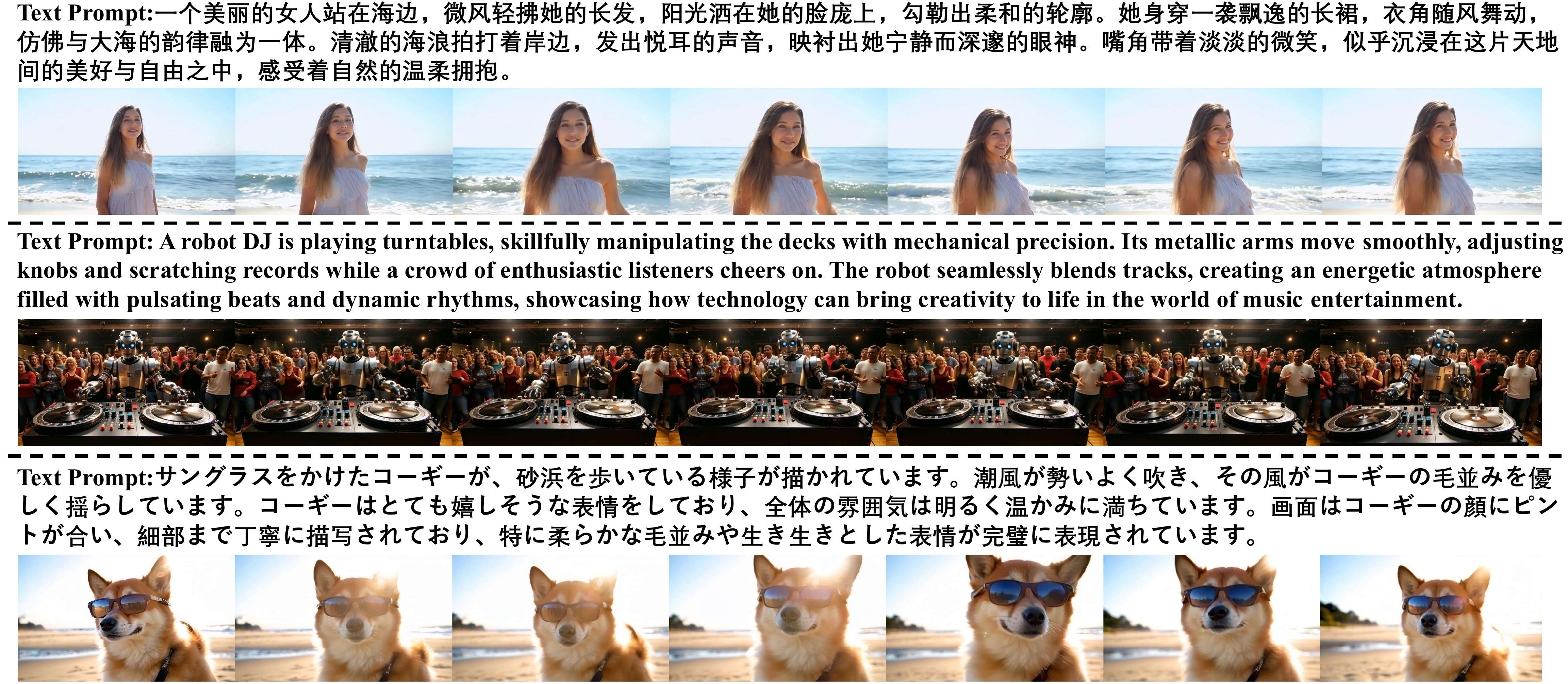}
  \caption{EasyAnimate can produce high-quality and coherent videos from multilingual text prompts.}
  \label{fig:teaser}
\end{teaserfigure}


\maketitle

\section{Introduction}
Artificial intelligence has broadened creative content generation across modalities.
Open-source projects like Stable Diffusion \cite{rombach2022high} have greatly advanced text-to-image generation.
Compared to images, video generation demands more computational resources and presents greater challenges due to temporal information.

Earlier video diffusion models are predominantly based on U-Net architectures~\cite{guo2024animatediff,chen2024videocrafter2,chen2023videocrafter1,wang2023modelscope,VideoFusion}. The recent introduction of Sora revolutionized the field with its diffusion transformer architecture~\cite{OpenAI2023VideoGeneration,zheng2024open,lin2024open,yang2025cogvideox,ma2025latte,polyak2024moviegencastmedia,hacohen2024ltx,kong2024hunyuanvideo}, marking a significant leap in video quality compared to previous models~\cite{OpenAI2023VideoGeneration}.
Despite these advancements, generating high resolution, high quality videos still faces substantial challenges.

The first challenge is the low training efficiency and slow inference speed. These are due to two factors: the high complexity of the transformer model and the uneven GPU utilization during training. 
Diffusion transformer-based models come with high computational costs, which grow quadratically with the sequence length~\cite{vaswani2017attention}. As videos naturally capture temporal information, they tend to produce longer sequences than images, thus exacerbating the problem. 
Some earlier works attempt to reduce complexity by employing spatial-temporal decoupled attention~\cite{zheng2024open,ma2025latte}. However, this method demonstrably compromises video generation quality, as it has a restricted receptive field and cannot capture large dynamic changes between frames. 
Some existing methods use 3D full attention to capture global video information~\cite{lin2024open,yang2025cogvideox,polyak2024moviegencastmedia,kong2024hunyuanvideo}. However, this approach demands substantial computational resources.
Inspired by recent progress in Large Language Models (LLMs)~\cite{beltagy2020longformer, jiang2023mistral7b}, we propose a novel multidirectional sliding window attention module to enlarge the receptive field across 3D dimensions.
Building on this module, we further propose Hybrid Windows Attention to strike a balance between computational efficiency and complexity. 
To address uneven GPU utilization, we design a token-based video training strategy that combines videos of varying resolutions and frame counts for joint training. 
By ensuring each sample has the same max token count, we balance GPU processing speed across samples, reducing idle time during training.

\begin{figure*}[h]
	\centering
	\includegraphics[width=0.98 \linewidth]{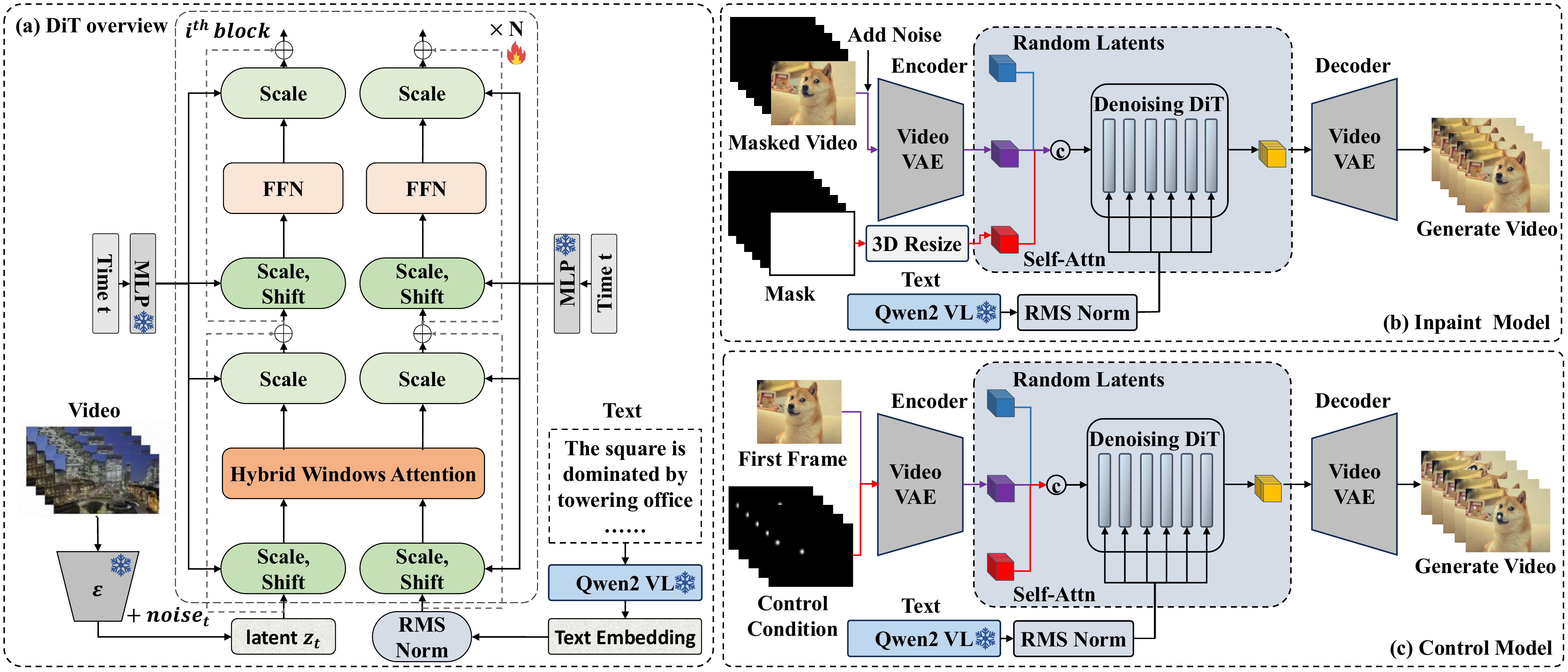}
	\caption{The detailed structure of the denoising diffusion transformer, inpaint model, and control model in EasyAnimate. }\label{fig:dit}
\end{figure*}

The second challenge is the suboptimal quality of video generation, which manifests in two areas: aesthetic divergence from human preferences and inaccurate adherence to text prompts.
To improve the human preference alignment of the model, we explore the reward backpropagation in EasyAnimate post-training, leveraging human preference models to steer the optimization process.
In this section, we experiment with different reward models and optimize their combinations. 
We find that combining different reward models can achieve superior performance. 
This strategy improves system performance and enhances the model’s capabilities, adaptability, and alignment with user preferences across diverse scenarios.
To better align text prompts with generated videos, we incorporate Multimodal Large Language Models (MLLMs) into video diffusion models to strengthen representation of detailed descriptions and complex object relationships.
Existing models typically use CLIP \cite{radford2021learning} or T5 \cite{raffel2020exploring} as text encoders, which restrict text length and hinder the understanding of detailed and complex scenes~\cite{zheng2024open,lin2024open,yang2025cogvideox,ma2025latte,hacohen2024ltx,genmo2024mochi}. 
MLLMs show strong performance on diverse text and vision-language tasks, offering enhanced text understanding for EasyAnimate.

Based on the above improvements, we develop a comprehensive framework for developing video diffusion models, named EasyAnimate. Our framework covers data preprocessing, variational autoencoder (VAE) training, diffusion transformer (DiT) model training, and post-training, which is illustrated in Figure \ref{fig:pipeline}.
Our contributions could be summarized as follows.

\textbf{(1)}~We propose Hybrid Windows Attention by interleaving multidirectional sliding window attention and full attention to boost the efficiency of video generation and training significantly.

\textbf{(2)}~We explore post-training with the Reward Backpropagation in the video diffusion transformers, which significantly improves the generated videos for better alignment with human preferences.

\textbf{(3)}~We propose an efficient and high-quality video generation framework called EasyAnimate. Within this framework, we incorporate improvements such as the Training with Token Length strategy and the use of MLLMs as the text encoder, thereby significantly enhancing both training efficiency and model performance.

\section{Related Work}

\subsection{Video Diffusion Model}
Recent studies on video generation have increasingly concentrated on diffusion models, which progressively enhance samples by iterative denoising from normal distributions.
Pioneering efforts \cite{guo2024animatediff,chen2024videocrafter2,chen2023videocrafter1,wang2023modelscope,VideoFusion} in video synthesis have utilized stable diffusion methods, with a focus on the U-Net architecture for the denoising process.
Recent studies suggest that diffusion transformer architecture significantly improves video generation capabilities~\cite{OpenAI2023VideoGeneration,zheng2024open,lin2024open,yang2025cogvideox,ma2025latte,polyak2024moviegencastmedia,hacohen2024ltx,kong2024hunyuanvideo} with improved text alignment and enhanced realism.
However, despite these advances, current video models still suffer from several notable limitations.
A challenging issue is the increased computational complexity of transformer architectures with longer sequences~\cite{vaswani2017attention}, resulting in slower generation and training speeds. 
Another point that can be improved is the reliance on CLIP and T5 as text encoders often results in limited language understanding capabilities~\cite{zheng2024open,lin2024open,yang2025cogvideox,ma2025latte,hacohen2024ltx,genmo2024mochi}.

To address these challenges, we design Hybrid Window Attention with a novel multidirectional sliding window attention module to improve efficiency. We further propose the Training with Token Length strategy to efficiently train videos of varying resolutions and frame counts. 
Additionally, we adopt MLLMs ($i.e.$, Qwen2-VL \cite{wang2024qwen2}) as the text encoder to improve text understanding capability.  

\subsection{Human Preference Alignment}
Diffusion models trained on large-scale web data with varying quality often produce visually unappealing or prompt-misaligned outputs. Previous works employ alignment training with reward models to address this issue. Existing alignment training methods fall into two categories: (1). Policy optimization \cite{black2023training, yang2024using, liu2024videodpo, xu2024visionreward}, which consider the diffusion sampling as a multi-step Markov decision process, using RL (or approximate) methods to optimize any black-box reward models; 
(2) Reward Backpropagation \cite{prabhudesai2023aligning, yuan2024instructvideo, clark2024directly, prabhudesai2024video}, which use differentiable reward models to directly guide diffusion sampling. These methods are typically more sample-efficient and effective than policy optimization approaches when a white-box reward model is available \cite{clark2024directly, prabhudesai2024video}. 
However, existing methods focus only on 2D VAE and U-Net based video diffusion models \cite{wang2023modelscope,blattmann2023stable} using DDPM sampling \cite{ho2020denoising}, which are now outperformed by diffusion transformer models employing 3D causal VAE and rectified flow sampling \cite{liu2023flow}. The use of reward backpropagation in such models remains unexplored and faces significant challenges. 
First, it involves multiple backpropagation steps through the 3D causal VAE and diffusion transformer, which have more parameters and larger activations, causing high GPU memory usage.
Second, We found directly applying DRaFT \cite{clark2024directly} or VADER \cite{prabhudesai2024video} to EasyAnimate results in training instability and reduces the dynamics of generated videos. 
In this work, we implement several crucial modifications to reward backpropagation to guarantee efficient training and convergence with rectified flow based video diffusion transformer models.

\section{Architecture}
EasyAnimate comprises a text encoder, a diffusion transformer and a video VAE. We first introduce our innovative Hybrid Windows Attention, followed by descriptions of the text encoder, diffusion transformer, and video VAE in sequence.

\begin{figure}[]
	\centering
	\includegraphics[width=0.98 \linewidth]{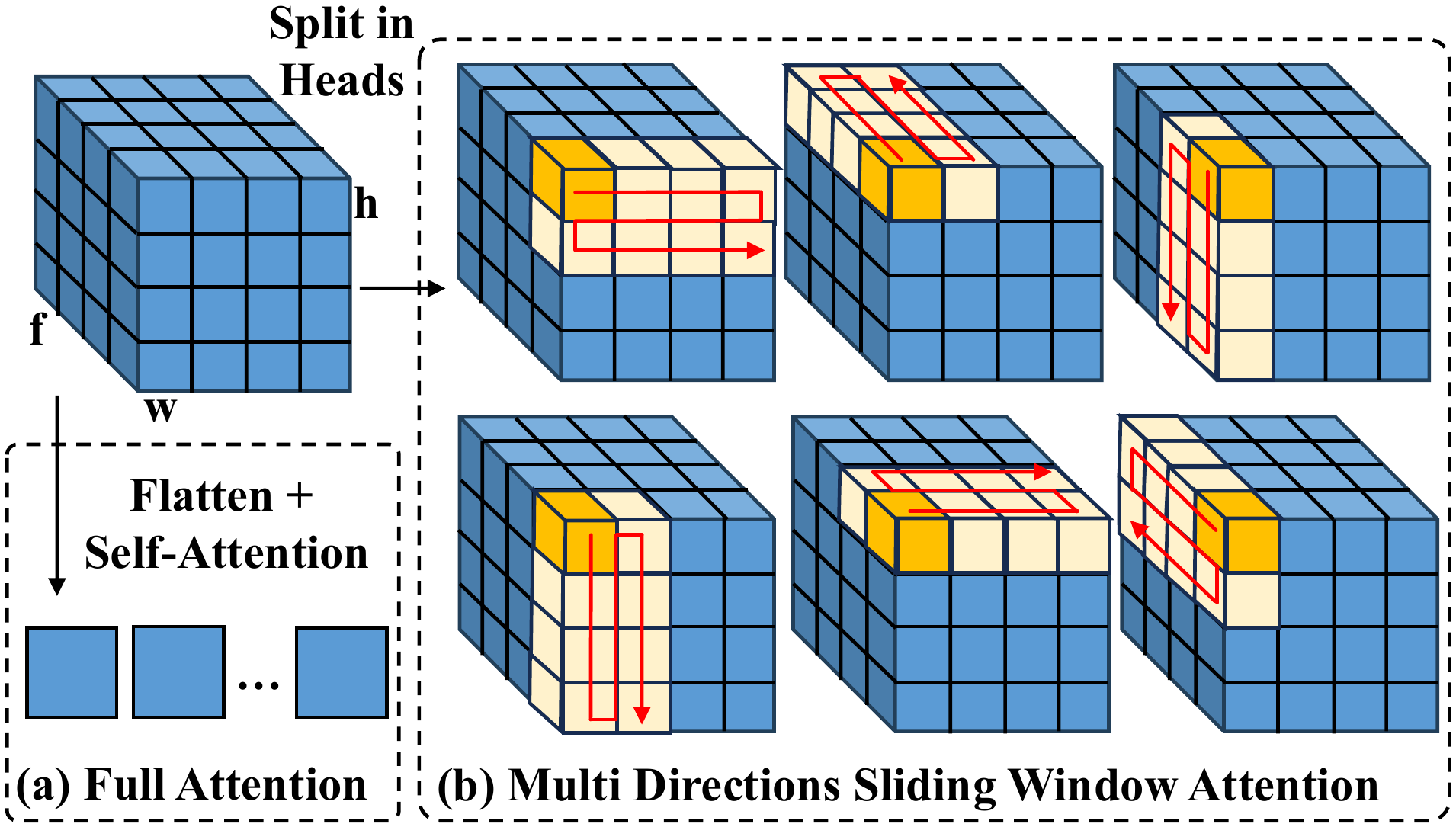}
	\caption{The details of full attention and multidirectional sliding window attention.}\label{fig:attention_compare}
\end{figure}

\subsection{Hybrid Window Attention}
Similar to existing diffusion transformer based video generation models, we initially explored 3D full attention. 
However, as video resolution and frame count increased, the computational cost grew quadratically.
For a model containing 12 billion parameters, producing a 1024×1024 video with 49 frames on a single A100 GPU required almost 30 minutes, posing a great challenge for many applications.
This underscores the necessity to reduce the model's computational costs.
Sliding window attention mechanism has been extensively utilized in large language models to reduce computational complexity~\cite{beltagy2020longformer,jiang2023mistral7b}. 
However, applying sliding window attention directly to video generation is inadequate, as existing window attention is single dimension, which fails to account for the 3D locality of video tokens and increases the risk of sudden changes.

\begin{algorithm}
\caption{Multidirectional Sliding Window Attention}
\label{alg:swa}
\begin{algorithmic}[t]
\STATE Split $Q, K, V$ into 6 head groups: $Qs, Ks, Vs$
\STATE Initialize $sliding\_dirs \leftarrow [fhw, fwh, hfw, hwf, wfh, whf]$
\FOR{$i=1$ to $6$}
    \STATE Rearrange $Qs[i]$, $Ks[i]$, $Vs[i]$ from $fhw$ to $sliding\_dirs[i]$
\ENDFOR
\STATE Concat $Qs$, $Ks$, $Vs$ into $Q$, $K$, $V$
\STATE $Q, K, V \leftarrow \text{WINDOW\_FLASH\_ATTENTION}(Q, K, V)$
\STATE Split $Q$, $K$, $V$ into 6 head groups: $Qs$, $Ks$, $Vs$
\FOR{$i=1$ to $6$}
    \STATE Rearrange $Qs[i]$, $Ks[i]$, $Vs[i]$ from $sliding\_dir[i]$ to $fhw$
\ENDFOR
\end{algorithmic}
\end{algorithm}

To address this issue, we propose a multidirectional sliding window attention module that partitions heads into groups, with each group performing sliding window attention in a different direction, as illustrated in Figure~\ref{fig:attention_compare}.
Compared to original one-dimensional sliding window attention, our multidirectional design greatly expands the model's 3D receptive field. It also enables efficient computation via standard multi-head attention libraries like FlashAttention \cite{dao2022flashattention}, as shown in Algorithm~\ref{alg:swa}. Alternative designs like spatial-temporal decoupled attention~\cite{zheng2024open} require multiple attention passes, while our approach needs only one, leading to higher efficiency.
Finally, we interleave 3D full attention with multidirectional sliding window attention, creating the Hybrid Window Attention model.
As shown in Table~\ref{tbl:speed_compare}, sliding window attention significantly reduces training and inference time, with this benefit becoming increasingly evident as sequence length grows.

\subsection{Text Encoder}
Existing text encoders such as CLIP and T5 generally suffers from limited text understanding ability, such as missing fine-grained details or misunderstanding complex object relationships. 
Moreover, the input length of the CLIP model is limited to 77 words, which is far from sufficient. 
Compared to CLIP and T5, MLLMs such as Qwen2-VL exhibit superior performance on various visual language understanding and reasoning tasks.
Unlike text-based models, MLLMs unify textual and visual tokens into a single representation space, which corresponds precisely to the task of video generation from text. We believe this is beneficial for optimizing diffusion models.
The Qwen2-VL-7B model is a leading example of MLLMs, achieving top performance among similarly scaled models. 
Thus, we select Qwen2-VL-7B as our text encoder, which also supports multilingual text inputs as an added benefit.

\begin{table}[t]
\centering	
\begin{tabular}	{l l l l}
    \toprule
        \textbf{Res.} & \textbf{Type} & \textbf{Train Latency ↓} & \textbf{Test Latency ↓} \\
         & & (s/Iter@1bs) & (s/Iter) \\
        \midrule	
        \textbf{768} & Full & 36.68 & 11.44 \\
         & Hybrid & 31.84~\textbf{(-13.19\%)} & 9.28~\textbf{(-18.89\%)} \\
        \midrule	
        \textbf{1024} & Full & 77.04 & 28.63 \\
         & Hybrid & 59.79~\textbf{(-22.39\%)} & 21.32~\textbf{(-25.53\%)} \\
    \bottomrule
\end{tabular}
\caption{
    Speed on A100 GPUs. \textbf{Hybrid} means Hybrid Windows Attention. \textbf{Full} means Full Attention.
}
\label{tbl:speed_compare}
\end{table}

We extract features from the penultimate hidden layer of Qwen2-VL.
The extracted textual features are then concatenated with video tokens into a single sequence to facilitate self-attention computation and further promote the alignment between multi-modal tokens. 
We notice that textual features often show a much larger L2 norm compared to video features, which start as white noise from a standard normal distribution. This discrepancy in L2 norm distribution leads to instabilities in optimizing diffusion models. To mitigate this issue, we apply RMSNorm \cite{zhang2019root} to the textual features. 
The normalized textual features are further transformed by a fully connected layer to reduce the discrepancy with video features. 

\subsection{Video Diffusion Transformer}

The video diffusion architecture is illustrated in Figure \ref{fig:dit}(a). In the model, we concatenate text and video embeddings for self-attention to promote alignment between visual and semantic information. 
However, significant disparities exist between the feature spaces of these two modalities, leading to potential discrepancies in the numerical scale of their embeddings.
To address this issue, we utilize MMDiT~\cite{sauer2024fast} as the foundational component of our model. Specifically, MMDiT incorporates distinct fully connected structures and feed-forward networks (FFNs) for each modality, thereby enhancing their alignment.
Following CogVideoX \cite{yang2025cogvideox}, we use 3D RoPE \cite{su2021enhanced} for positional embeddings by applying 1D RoPE to each spatial dimension and allocating 3/8, 3/8, and 2/8 of the hidden channels, which are then concatenated to form the final 3D RoPE encoding.
Initial experiments showed rectified flow loss outperforms DDPM loss, so we adopt it in our experiments.

We also build an inpaint model that reconstructs targeted regions by incorporating reference images and masks, enabling video generation from start and end frames and supporting video editing, as shown in Figure~\ref{fig:dit}(b).
Moreover, we train a multifunctional control model using conditions like trajectory, OpenPose, scribble, canny, MLSD, HED, and depth, with details presented in Figure~\ref{fig:dit}(c).

\subsection{3D Causal VAE}
To mitigate the computational complexity stemming from the 3D nature of video data, we use a 3D causal VAE to compress the videos across both spatial and temporal dimensions. Despite its efficiency gains, the VAE itself demands significant computational resources. 
The causal property of the 3D VAE allows us to cache the previous latent state and connect it with the next frame for processing.
We apply spatial and temporal slicing to the VAE, greatly reducing memory use during long, high-resolution video decoding.
During VAE training, we sample frames at varying intervals to improve robustness in cross-frame encoding and decoding.
Following MovieGen \cite{polyak2024moviegencastmedia}, we add a loss term to penalize latent encodings, reducing speckle artifacts during pixel-space video decoding.

\section{Model Training}

The model training is divided into four steps: data preprocessing, VAE training, DiT training, and post-training. We provided a detailed explanation excluding the VAE training.

\subsection{Data Curation}
We collect raw videos from public datasets including Panda-70M \cite{chen2024panda}, InternVid \cite{wang2024internvid}, MiraData \cite{ju2025miradata} and Pexels \cite{pexels}, as well as from internal sources. To construct a high-quality video dataset for training, we use a data preprocessing pipeline, shown in Figure \ref{fig:pipeline}, consisting of three stages: video splitting, filtering, and captioning.

\noindent \textbf{Video Splitting}: We detect scene changes with PySceneDetect \cite{pyscenedetect} and split videos into single-shot clips using FFmpeg \cite{ffmpeg}, following Panda-70M \cite{chen2024panda}. Video clips under 3 seconds are discarded, while those over 10 seconds are recursively split.

We identified some video clips still containing scene transitions: 1. starting with frames from the previous scene; 2. ending with frames from the next scene; 3. fade-ins or fade-outs. To mitigate their impact on temporal consistency, we discarded the beginning and ending frames. Next, we extracted I-frames near scene changes and combined them with the first and last frame to compute semantic consistency using CLIP and DINO \cite{oquab2024dinov2}.

\noindent \textbf{Video Filtering}: 
Based on the video clips obtained in the previous stage, we sequentially filter out low-quality data to avoid harming the model, based on three dimensions:

\textbf{I. Aesthetic Score}: 
To filter out video clips with poorly aesthetic and visually unappealing content, we calculate the average aesthetic score of uniformly sampled frames using the SigLIP-based aesthetic score predictor \cite{aesthetic25}, which outperforms the original LAION CLIP-based prediction model \cite{schuhmann2022laion} in our evaluation.

\textbf{II. Text Score}: 
To filter out video clips containing excessive text ($e.g.$, subtitles), we apply the text detection model CRAFT \cite{baek2019character} to calculate the average text area proportion of uniformly sampled frames as the text score.

\textbf{III. Motion Score}: To filter clips with low motion ($e.g.$, static images) or extremely dynamics ($e.g.$, slideshow), we use the Farneback algorithm \cite{farneback2003two} in OpenCV \cite{opencv} to compute the average optical flow between frames as the motion score. Additionally, we train a classifier to detect camera shake in video clips, which motion scores often fail to capture.

Given the above three dimensions, we progressively increase the corresponding filtering thresholds along with resolution to obtain datasets for different training stages, as shown in Table \ref{tbl:dataset_filter}.

\begin{table}[!t]
\centering	
\begin{tabular}	{l | c c c}
    \toprule
        \textbf{Stage} &
        \textbf{Pretrain} &
        \textbf{Pretrain-HR} &
        \textbf{Finetune} \\
        \midrule	
        \textbf{Video Clips} & 33.72M & 25.10M & 0.47M \\
        \textbf{Image Clips} & 2.87M & 2.87M & 0.04M \\
        \textbf{Video Source} & ALL & ALL & HQ \\
        \midrule
        \textbf{Res. Thr.} & - & 512 & 720 \\
        \textbf{Motion Thr.} & 0.50 & 0.50 & 2.00 \\
        \textbf{Aesthetic Thr.} & 4.00 & 4.00 & 4.50 \\
        \textbf{Text Thr.} &  0.02 & 0.02 & 0.02 \\        
    \bottomrule
\end{tabular}
\caption{
    The dataset after filtering with different constraints. The \textbf{Thr.} means the threshold. The \textbf{Res.} means the resolution. The \textbf{HQ} means high quality videos. 
}
\label{tbl:dataset_filter}
\end{table}		

\noindent \textbf{Video Captioning}: 
Recent studies \cite{betker2023improving, OpenAI2023VideoGeneration} highlight the value of dense and short captions in visual generation. We employ InternVL2-40B \cite{chen2024internvl} to generate dense captions for video clips and refine them with LLama-3-70B \cite{dubey2024llama} to remove subjectivity and enhance suitability as training prompts. Additionally, LLama-3-70B summarizes a subset of dense captions into short captions.
To address MLLMs' inability to describe camera movements ($e.g.$, tilt, dolly), we trained a specialized classification model and integrated detected movements into video captions. Additionally, we used VideoCLIP-XL-v2 \cite{wang2024videoclip} to compute caption-video similarity, ensuring alignment and enhancing prompt-following capability in training datasets.

We curate approximately 34M video-text and 3M image-text pairs for joint image-video training, using aesthetic filtering on JourneyDB \cite{sun2023journeydb} and caption annotations from ALLaVa \cite{chen2024allava}.

\subsection{DiT Training}

\begin{figure}[t]
	\centering
	\includegraphics[width=0.98 \linewidth]{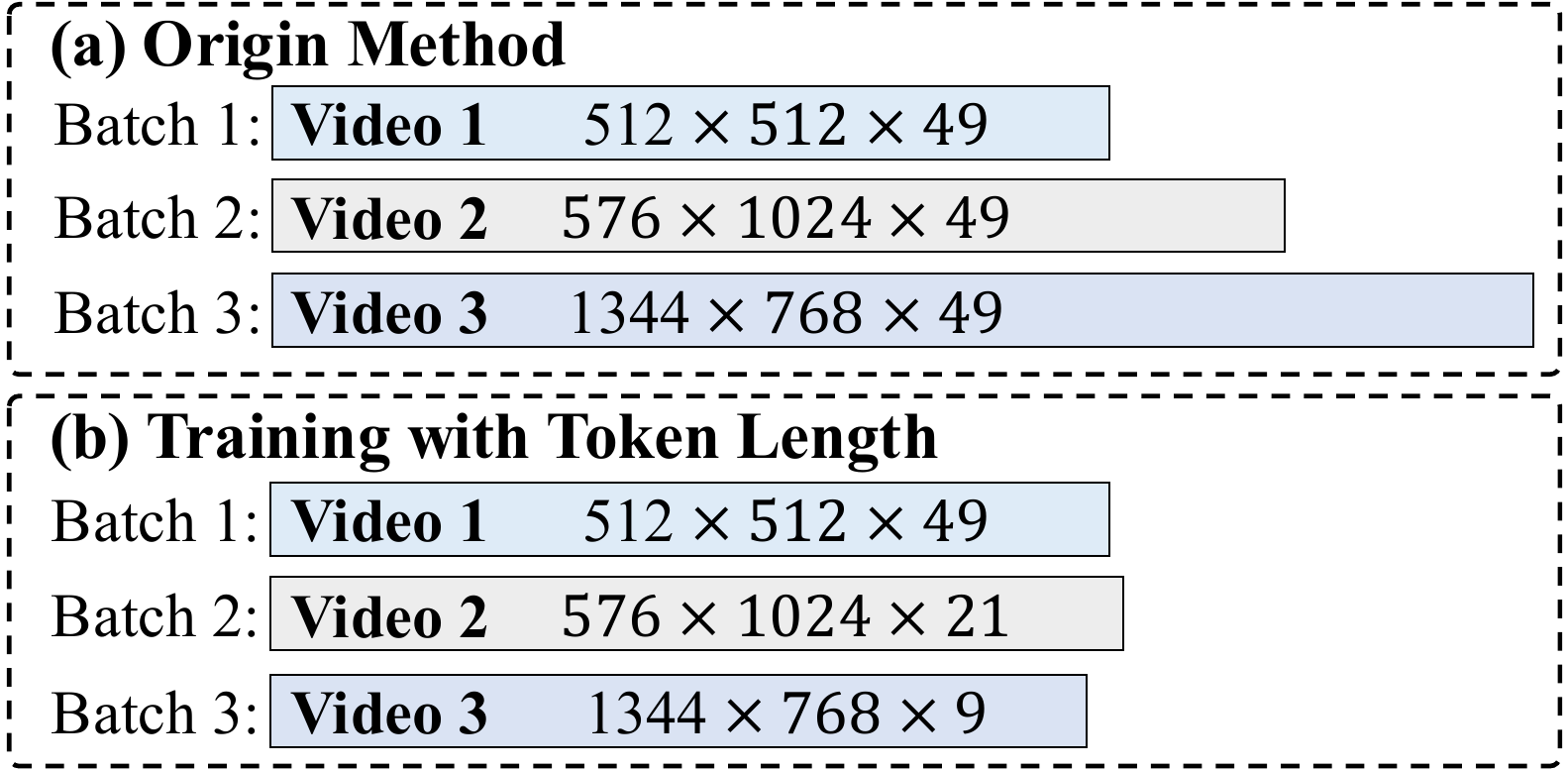}
	\caption{Illustration of Training with Token Length. We train videos with similar token lengths in one step. }\label{fig:bs_compare}
\end{figure}

\noindent \textbf{Training with Token Length:}~We design a new video training strategy based on token length. As a key engineering optimization, it allows the model to adapt to different resolutions and frame counts while improving training efficiency. 
The main factor influencing the training speed of diffusion transformers is sequence length, which is further dictated by the combination of video resolution and video length. 
As the model is trained on GPU cluster, the workload on different GPUs could be seriously unbalanced under naive settings, as indicated by Figure \ref{fig:bs_compare}(a). 
To balance the workload across different GPUs, we selected samples with similar token lengths at each training phase. 
As shown in Figure \ref{fig:bs_compare}(b), a 49-frame video at $512\times512$ resolution and a 21-frame video at $768\times768$ resolution have comparable sequence lengths; therefore, they can be jointly trained in the same phase.
We measure the efficiency of the training strategies by the total number of tokens trained per iteration. Our method demonstrates an improvement of 120.91\% compared to the naive method, as shown in Table \ref{tbl:token}.

\noindent \textbf{Progressive Training:}~
The EasyAnimate follows a multi-stage training process. 
Following PixArt~\cite{chen2023pixartalpha}, EasyAnimate adopts a progressive training strategy, moving from lower to higher resolutions. 
Unlike PixArt, our approach involves the utilization of reward models during the post-training phase. 
The training stages before post-training are outlined as follows.

\textbf{I. VAE-adapt}: Aligns DiT parameters with VAE using SAM~\cite{kirillov2023segment} image data.

\textbf{II. Pretraining}: Pretraining starts with an initially filtered dataset (\textbf{Pretrain} in Table \ref{tbl:dataset_filter}), using a token length of $256\times256\times49$. Subsequently, continued pretraining employs a resolution-filtered dataset (\textbf{Pretrain-HR} in Table \ref{tbl:dataset_filter}) with a token length of $512\times512\times49$.

\textbf{III. Finetune}: Finetuning model's image-to-video capabilities by a finely filtered dataset (Finetune in Table \ref{tbl:dataset_filter}), initially with a token length of $512\times512\times49$, followed by a token length of $1024\times1024\times49$. 

\subsection{Post Training with Reward Backpropagation}
\begin{table}[t]
\centering	
\begin{tabular}	{l | c c}
    \toprule
        \textbf{Methods} &
        \textbf{Origin} & \textbf{TTL} \\
        \midrule	
        \textbf{Tokens/Iter@256bs ↑} & 6.17m & 13.63m \textbf{(+120.91\%)} \\
    \bottomrule
\end{tabular}
\caption{Comparison of "Origin Method" and "Training with Token Length" (TTL) in terms of the number of tokens trained per iteration (Batch Size: 256, Resolution: 1024$\times$1024, Frame Count: 49).}
\label{tbl:token}
\vspace{-2.0em}
\end{table}		
\begin{table*}[!t]
\centering	
\begin{tabular}	{l | c c c c c c c c}
    \toprule
        Models &
        \textbf{Total} &
        \textbf{Quality} &
        \textbf{Semantic} &
        \textbf{Aesthetic} & 
        \textbf{Subject} & 
        \textbf{Spatial} & 
        \textbf{Object} & 
        \textbf{Scene} \\
        ~ &
        \textbf{Score} &
        \textbf{Score} &
        \textbf{Score} &
        \textbf{Quality} & 
        \textbf{Consistency} & 
        \textbf{Relationship} & 
        \textbf{Class} & 
        \textbf{~} \\ 
        \midrule
        \textbf{AnimateDiff-V2} & 80.27 & 82.90 & 69.75 & 67.16 & 95.30 & 34.60 & 90.90 & 50.19\\
        \textbf{VideoCrafter-2.0} & 80.44 &  82.20 &  73.42 & 63.13 & 96.85 & 34.60 & 92.55 & 42.44\\
        \textbf{OpenSora V1.2} & 79.76 &  81.35 & 73.39 & 56.85 & 96.75 & 68.56 & 82.22 & 50.19\\
        \textbf{OpenSoraPlan V1.3} & 77.23 & 80.14 & 65.62 & 60.42 & 97.79 & 51.61 & 85.56 & 36.73\\
        \textbf{CogVideoX1.5-5B} & 82.17 & 82.78 & 79.76 & 62.79 & 96.87 & 80.25 & 87.47 & 52.91\\
        \textbf{CogVideoX-5B} & 81.61 & 82.75 & 77.04 & 61.98 & 96.23 & 66.35 & 85.23 & 53.20\\
        \textbf{HunyuanVideo} & 83.24 & 85.09 & 75.82 & 60.36 & 97.37 & 68.68 & 86.10 & 53.88\\
        \textbf{Jimeng} $\ddag$ & 81.97 & 83.29 & 76.69 & 68.80 & 97.25 & \textbf{77.45} & 89.62 & 44.94\\
        \textbf{Vidu} $\ddag$ & 81.89 & 83.85 & 74.04 &  60.87 & 94.63 & 66.18 & 88.43 & 46.07\\
        \textbf{Gen-3} $\ddag$ & 82.32 & 84.11 & 75.17 &  63.34 & 97.10 & 65.09 & 87.81 & 54.57\\
        \textbf{MiniMax-01} $\ddag$ & 83.41 & 84.85 & 77.65 &  63.03 & 97.51 & 75.50 & 87.83 & 50.68\\
        \textbf{Sora} $\ddag$ & \textbf{84.28} & \textbf{85.51} & \textbf{79.35} & 63.46 & 96.23 & 74.29 & \textbf{93.93} & \textbf{56.95}\\
        \midrule	
        \textbf{EasyAnimate} & 83.42 & 85.03 & 77.01 & 69.48 & \textbf{98.00} & 76.11 & 89.57 & 54.31\\
        \textbf{EasyAnimate-Hybrid} & 83.27 & 84.70 & 77.54 & \textbf{70.64} & 97.76 & 77.32 & 92.24 & 56.10\\
    \bottomrule
\end{tabular}
\caption{
    Comparison of EasyAnimate with SOTA models on VBench~\cite{huang2024vbench} (up to the submitted time of EasyAnimate, $i.e.$, 2025-01-22.). EasyAnimate-Hybrid refers to EasyAnimate with Hybrid Windows Attention. $\ddag$ indicates a closed-source model. 
}
\label{tbl:evaluation_result}
\end{table*}		
After pretraining on large scaled text-video paired datasets, the model could generate videos according to textual prompts. Nevertheless, the generated videos might fall short of human performance due to the vast expressive space inherent in videos. A close examination of the initial generation results show that certain detailed textual descriptions are overlooked, and some of the videos could not achieve the aesthetic level of cinematic quality.
To further enhance the quality of generated videos, we adopt reward backpropagation \cite{clark2024directly,prabhudesai2024video} with LoRA \cite{hu2022lora} to fine-tune the DiT model for better alignment with human preferences. Given a differentiable reward model $R$ \cite{aesthetic25,wu2023human,zhang2024learning}, reward backpropagation aims to optimize the DiT parameters $\theta$ so that videos generated by the sampling process maximize empirical reward. The objective can be formulated as:

\begin{equation}
    L(\mathbf{\theta}) = -\frac{1}{|\mathcal{P}|} \sum_{\mathbf{c} \in \mathcal{P}} R(\text{sample}(\theta, \mathbf{c}, \mathbf{x}_{T}), \mathbf{c}) \label{eq:reward_backpropagation}
\end{equation}

where $\text{sample}(\theta, \mathbf{c}, \mathbf{x}_{T}), \mathbf{c})$ refers to the sampling process from time $t = T \to 0$ with condition $\mathbf{c}$, $\mathcal{P}$ refers to the prompt training dataset.

In fact, not all denoising steps in the sampling chain require backpropagation. To save GPU memory and reduce computation, previous works \cite{clark2024directly,prabhudesai2024video} only optimize the last step ($i.e., K \to 0$, where $K=1$), while the beginning $T \to K$ steps are detached from the computation graph. However, we find optimizing only the last step in EasyAnimate is far from sufficient: the convergence speed is slow and not stable. A detailed analysis of the training process reveals that the gradient norm is considerably smaller when employing a rectified flow-based probability path compared to a DDPM-based probability path.
Detailed comparisons of $K$ are shown in Section \ref{sec:ablation_k}. As EasyAnimate utilizes flow-matching sampling in both training and inference, we set $K=10$ in EasyAnimate.

Besides, VADER calculates the reward on multiple uniformly sampled frames with an image-based reward model. However, we found that calculating rewards across multiple frames not only consumes more GPU memory but also impairs the dynamics and  generalization of the generated videos. Detailed results are shown in Section \ref{sec:ablation_f}. Thus, we set $F=1$ in EasyAnimate.

\section{Experiment}
\subsection{Evaluation}

\noindent \textbf{Automated Evaluation}: To comprehensively evaluate the performance of text-to-video generation models, we employ a series of metrics on the VBench~\cite{huang2024vbench}.
We primarily focus on the Total Score, Quality Score, and Semantic Score. The Total Score is the overall score from VBench, the Quality Score emphasizes visual quality, and the Semantic Score focuses on semantic information.
We compare the performance of EasyAnimate with other models in Table~\ref{tbl:evaluation_result}.
EasyAnimate achieves the best performance across multiple metrics and demonstrates competitive results. 
%
Particularly in the aesthetic metrics, guided by human preference models, EasyAnimate's generated results exhibit excellent aesthetic quality.
These findings show that EasyAnimate excels in both video generation quality and prompt semantics interpretation, accurately capturing object relationships.

\begin{table}[t]
\centering	
\begin{tabular}	{l | c c c}
    \toprule
        \textbf{Models } &
        \textbf{Quality} &
        \textbf{Semantic} &
        \textbf{Physics} \\
        \midrule	
        \textbf{CogVideoX} & 17.08\% & 18.63\% & 21.73\% \\
        \textbf{HunyuanVideo} & 32.61\% & 37.28\% & 33.24\% \\
        \textbf{EasyAnimate} & \textbf{50.31\%} & \textbf{44.09\%} & \textbf{45.03\%} \\
    \bottomrule
\end{tabular}
\caption{
    Win rates of different models and different aspects in human evaluation.
}
\label{tbl:human_evaluation}
\end{table}		
\noindent \textbf{Human Evaluation}: 
Besides automatic evaluation with VBench, we conduct a comparative analysis involving human evaluations on HunyuanVideo \cite{kong2024hunyuanvideo}, CogVideoX, and EasyAnimate. 
We randomly selected 100 prompts from T2V-CompBench \cite{sun2025t2v}, which cover various aspects, was provided to human evaluators. To ensure unbiased assessments, the videos were shuffled for a process of blind evaluation.
The quality of the generated videos was evaluated based on three key criteria: perceptual quality, text-video consistency, and adherence to physical laws. 
As shown in Table~\ref{tbl:human_evaluation}, the results demonstrate that EasyAnimate achieved the highest preference from human evaluators across all categories. 

\subsection{Ablation Study}

\noindent \textbf{Ablation of different text encoders}:
In this section, we analyze the impact of different text encoders on performance, as shown in Table~\ref{tbl:text_encoders}.
We first implement a dual encoder combining CLIP and T5, following SD3~\cite{esser2024scaling}.
However, CLIP limits text to 77 tokens, and T5's ability to understand nuanced text is suboptimal.
To address this, we adopt Qwen2-VL as the text encoder.
VBench results show that Qwen2-VL significantly improves overall performance.

\begin{table}[t]
\centering	
\begin{tabular}	{l | c c c}
    \toprule
        \textbf{Text Encoders} &
        \textbf{Total} &
        \textbf{Quality} &
        \textbf{Semantic} \\
        \midrule	
        \textbf{T5 + CLIP} & 80.42\% & 82.56\% & 71.85\% \\
        \textbf{Qwen2 VL} & \textbf{81.57\%} & \textbf{83.52\%} & \textbf{73.76\%} \\
    \bottomrule
\end{tabular}
\caption{
    The impact of text encoders by scores in VBench. 
}
\label{tbl:text_encoders}
\end{table}		

\noindent \textbf{Ablation of Hybrid Windows Attention}: 
In this study, we conduct ablation studies across three key dimensions: (1) the position of window attention within the network, (2) the window size, and (3) the number of directions. 
We selected 1,000 videos from the WebVid validation set to calculate the FVD score. 
We first apply multidirectional sliding window attention in shallow (1-24), middle (12-36), and deep (24-48) layers.
Table~\ref{tbl:positions} shows that using it in middle layers hurts performance the least. 
We hypothesize that not all layers require global information. Using window attention in middle layers allows the model to inherit global context from earlier full-attention layers and maintain stability via later ones.
The Table~\ref{tbl:windows_size} shows that decreasing the window size worsens FVD without notable speed gains, while increasing it offers no substantial FVD boost. The current setting balances speed and quality. 
In addition, we tested window attention with 1, 3, and 6 directions, with 6-directional performing best at an FVD score of 352.3, compared to 373.6 for 3-directional and 408.1 for 1-directional. 

\begin{table}[t]
\centering	
\begin{tabular}	{l | c c c c}
    \toprule
        \textbf{Positions} &
        \textbf{N/A}&
        \textbf{Shallow}&
        \textbf{Middle}&
        \textbf{Deep} \\
        \midrule	
        \textbf{FVD Score ↓} & 364.9 & 459.7 & \textbf{352.3} & 353.6 \\
    \bottomrule
\end{tabular}
\caption{
     The impact of different positions of multidirectional sliding window attention in EasyAnimate.
}
\label{tbl:positions}
\end{table}		
\begin{table}[t]
\centering	
\begin{tabular}	{l | c c c c}
    \toprule
        \textbf{Window Size} &
        \textbf{H*W/8} &
        \textbf{H*W/2} &
        \textbf{H*W} &
        \textbf{H*W*2} \\
        \midrule
        \textbf{FVD Score ↓} & 557.0 & 385.5 & 352.3 & \textbf{348.3} \\
        \textbf{Time (s/Iter) ↓} & \textbf{19.81} & 20.43 & 21.32 & 22.73\\
    \bottomrule
\end{tabular}
\caption{
     The impact of different windows sizes. \textbf{H} means height of the feature. \textbf{W} means width of the feature. 
}
\label{tbl:windows_size}
\end{table}

\begin{figure}[t]
    \centering
    \includegraphics[width=0.98 \linewidth]{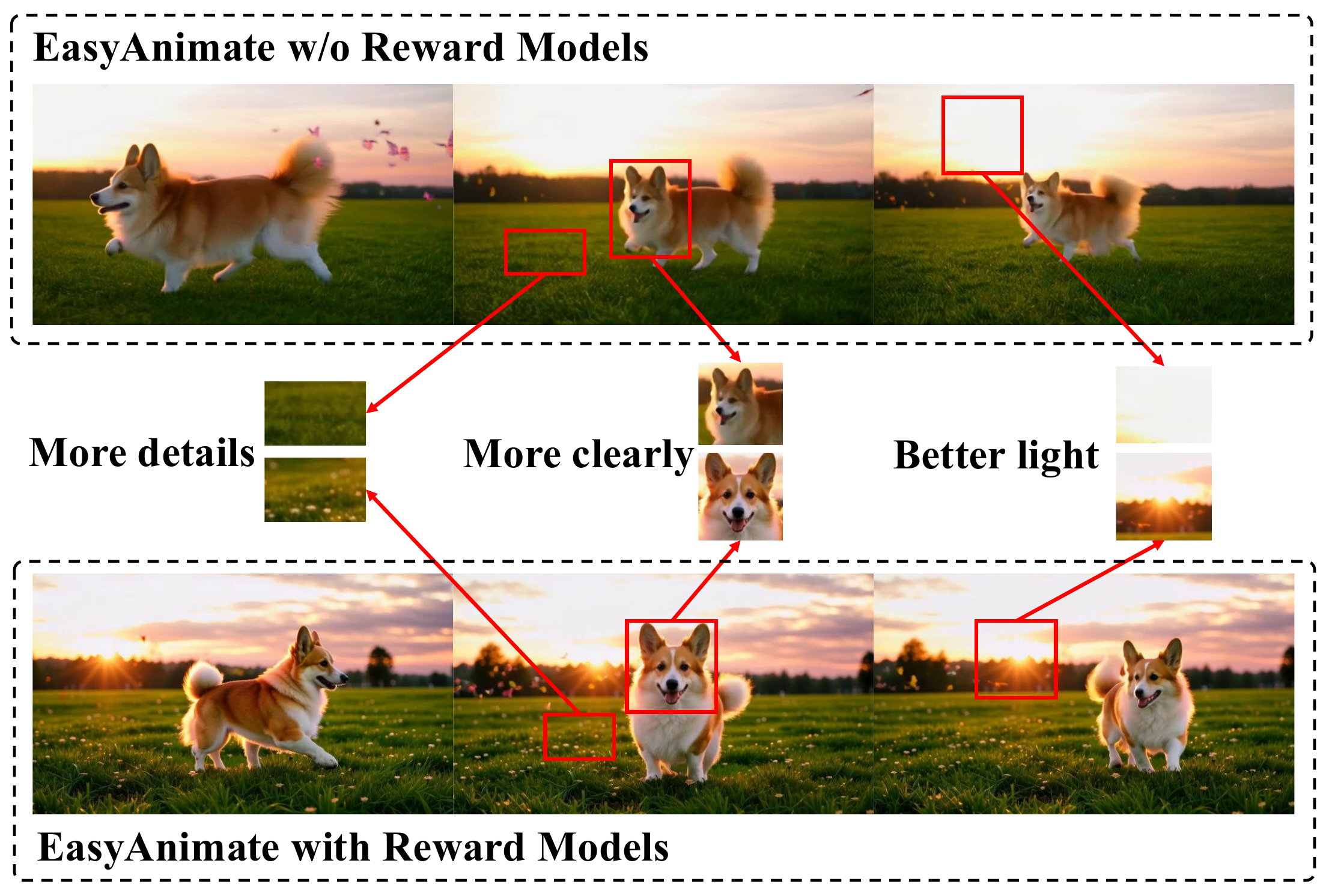}
    \caption{Comparison of evaluation results between EasyAnimate with and without reward models.}
    \label{fig:reward_models_compare}
\end{figure}

\noindent \textbf{Ablation of different reward models}: In this study, we explore reward backpropagation to optimize generated videos for better alignment with human preferences. 
We first explore the impact of distinct reward models, specifically the SigLIP-based aesthetic score predictor \cite{aesthetic25}, MPS \cite{zhang2024learning} and HPSv2.1 \cite{wu2023human}, on the model's performance. 
Our results show that both MPS and HPSv2.1 significantly improve the VBench composite score of generated videos.
We observe further improvements in both quality and semantic scores.
In further experiments, we explore the performance of integrating different reward models. 
The integration of HPSv2.1 with MPS yields the optimal performance.
Figure~\ref{fig:reward_models_compare} compares EasyAnimate outputs with and without reward feedback, using the same prompt.
The reward-optimized model produces clearer, more textured visuals and richer details in the generated results. 
In conclusion, the improved model better aligns with human preferences.

\begin{table}[t]
\centering	
\begin{tabular}	{l | c c c}
    \toprule
        \textbf{Reward models} &
        \textbf{Total} &
        \textbf{Quality} &
        \textbf{Semantic} \\
        \midrule	
        \textbf{N/A} & 81.57\% & 83.52\% & 73.76\% \\
        \textbf{Aesthetic} & 81.72\% & 83.60\% & 74.19\% \\
        \textbf{MPS} & 82.36\% & 84.07\% & 75.52\% \\
        \textbf{HPSv2} & 83.26\% & 84.87\% & 76.79\% \\
        \textbf{HPSv2 + Aesthetic} & 83.24\% & 84.91\% & 76.55\% \\
        \textbf{HPSv2 + MPS} & \textbf{83.42\%} & \textbf{85.03\%} & \textbf{77.01\%} \\
    \bottomrule
\end{tabular}
\caption{
    We explore the impact of different reward models by scores in VBench. The \textbf{Total} means Total Score. The \textbf{Quality} means Quality Score. The \textbf{Semantic} means Semantic Score. 
}
\label{tbl:reward_models}
\end{table}		

\begin{figure}[t]
	\centering
	\includegraphics[width=0.98 \linewidth]{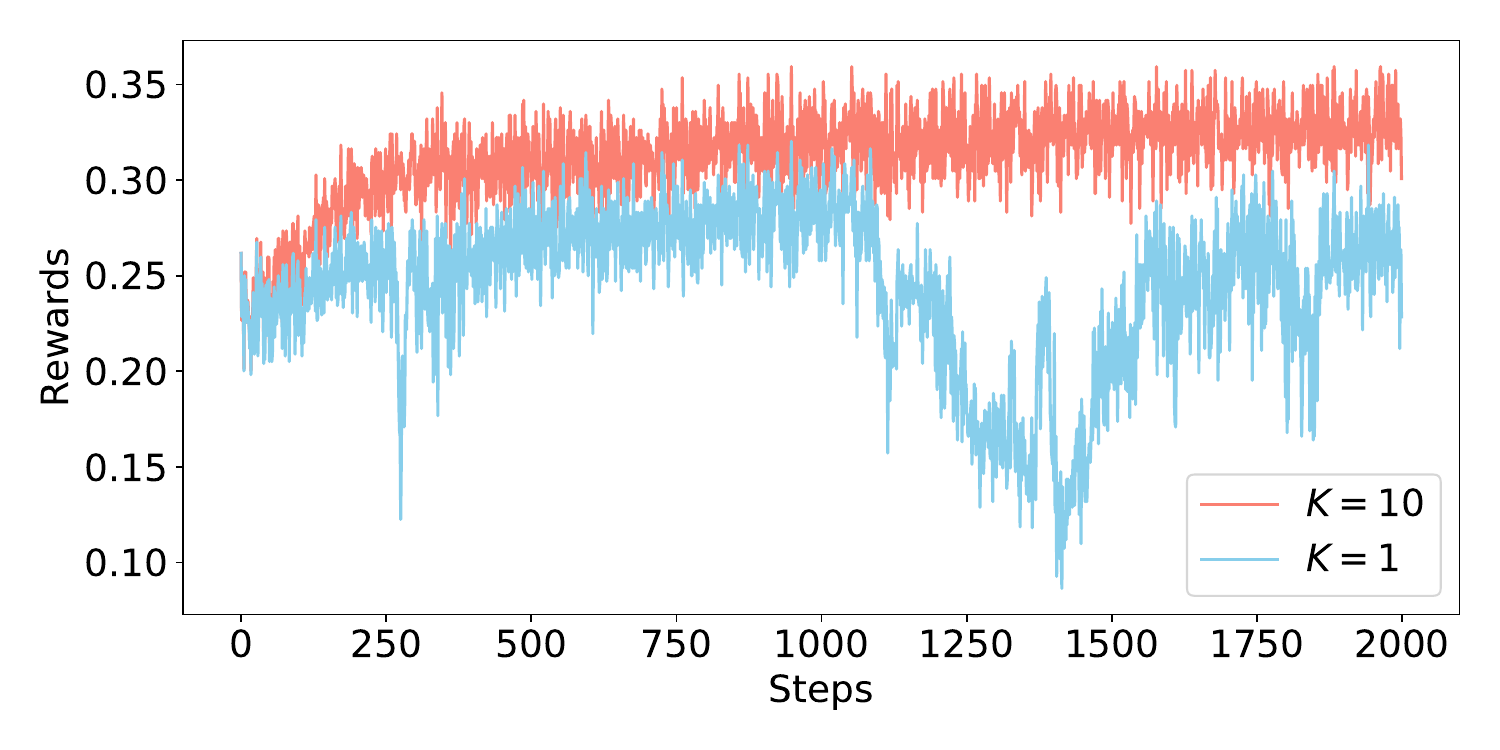}
	\caption{The impact of reward backpropagation steps $K$. }\label{fig:ablation_reward_k}
\end{figure}

\noindent \textbf{Ablation of Backpropagation Steps $K$} \label{sec:ablation_k}:
We conduct an ablation study on the selection of $K$ in EasyAnimate with the HPSv2.1 reward model. As shown in Figure \ref{fig:ablation_reward_k}, it can be seen that performing reward backpropagation only at the final step of the denoising process is not sufficiently stable, as evidenced by the sudden drop in training rewards. This may be due to the gradient norm being much smaller for $K=1$ compared to $K=10$.

\begin{figure}[t]
    \centering
    \includegraphics[width=0.98 \linewidth]{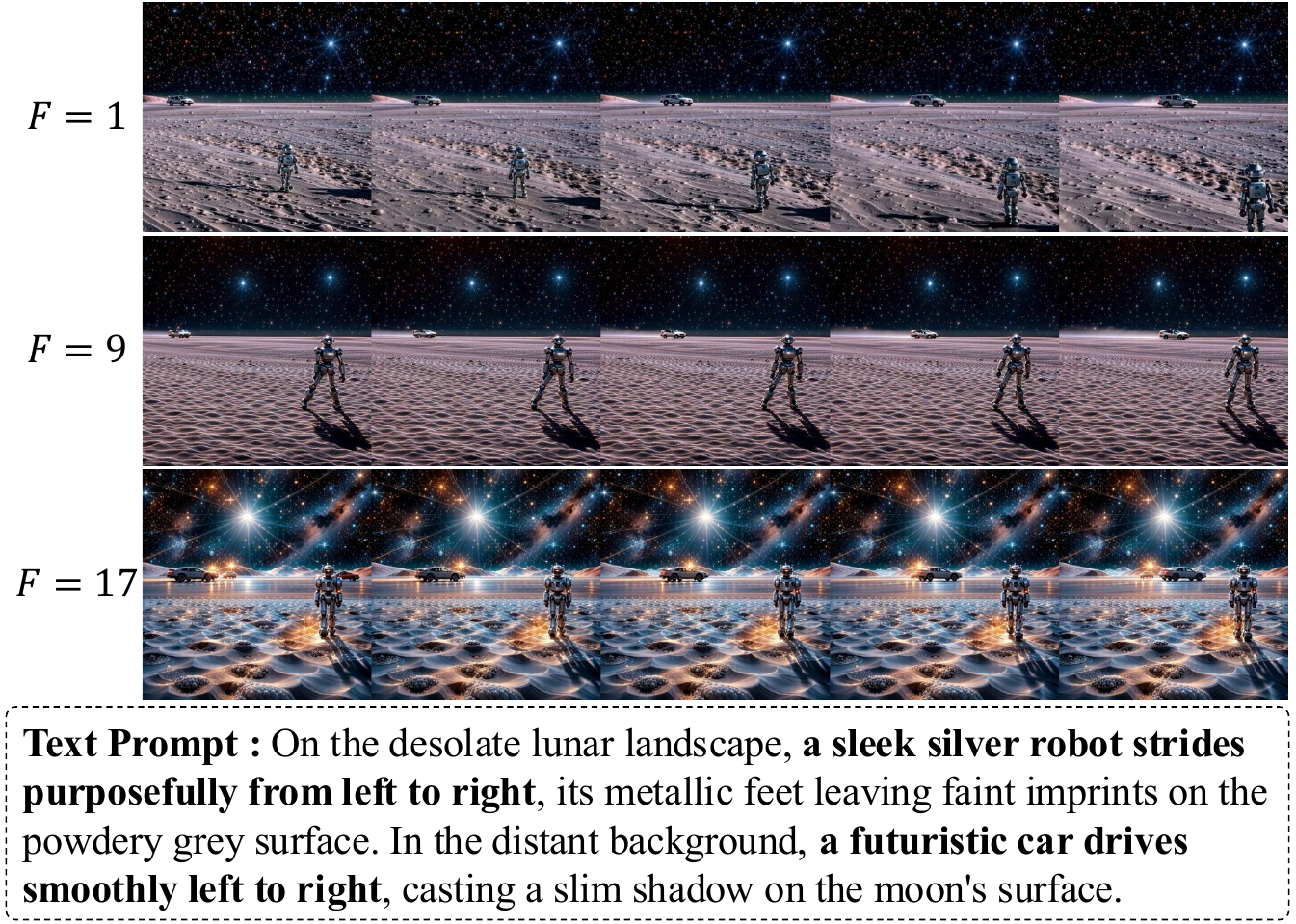}
    \caption{The impact of reward decoding frames $F$. } \label{fig:ablation_reward_f}
\end{figure}

\noindent \textbf{Ablation of Decoding Frames $F$} \label{sec:ablation_f}: We conduct an ablation study on the selection of $F$ in EasyAnimate with the HPSv2.1 reward model. It can be seen that extracting multiple frames for reward backpropagation impairs the dynamics of the robot movement in Figure \ref{fig:ablation_reward_f}. Furthermore, when $F$ is too large ($e.g., F=17$), the training process is more prone to reward hacking, which can be observed by artifacts in the video background from Figure \ref{fig:ablation_reward_f}. We speculate that this is due to the use of an image-based reward model, where extracting multiple frames may lead to conflict optimization directions between frames. Thus, setting $F=1$ is sufficient to ensure training convergence and generalization in video generation with the 3D Causal VAE, which can refer to the first frame to decode the remaining video frames.

\section{Limitations}
EasyAnimate has limitations in color accuracy and dynamic degree, likely due to dataset processing issues. For example, the model may generate a green apple and a green cup when asked for a green apple and a yellow cup, significantly affecting visual fidelity. Additionally, it currently only supports generating videos up to 5 seconds long, limiting its applicability for longer-duration tasks.

\section{Conclusion}
In this paper, we present EasyAnimate, a versatile video generation framework leveraging transformer-based architecture to produce coherent videos. 
To address the computational demands of long video sequences, we introduce Hybrid Windows Attention, based on a multidirectional sliding window module, which reduces complexity while improving temporal and spatial dependency modeling.
To boost video generation performance and improve alignment with human preferences, we refine EasyAnimate using reward models.
Additionally, we propose a training strategy to improve efficiency when training videos of varying resolutions and frame counts.
To further improve text understanding, we adopt MLLMs as the text encoder, which also enables multilingual support.
Experiments show SOTA performance on the video evaluation leaderboard, highlighting EasyAnimate's advancements in video generation.
\bibliographystyle{ACM-Reference-Format}
\bibliography{sample-base}

\end{document}